\documentclass[letterpaper, 10 pt, conference]{ieeeconf}  

\IEEEoverridecommandlockouts                              

\usepackage[T1]{fontenc}
\usepackage{marvosym}
\usepackage[utf8]{inputenc}
\usepackage{graphicx}
\usepackage{hyperref}
\usepackage{CJKutf8}
\usepackage{svg} 
\usepackage{pdfpages}
\usepackage{subfigure}
\usepackage{tabularx}
\usepackage{booktabs}
\usepackage{amsmath}
\usepackage{amssymb}
\usepackage{mathtools}
\usepackage{bm}
\usepackage{amssymb} 
\usepackage{CJKutf8}
\usepackage{booktabs}
\usepackage{multirow}
\usepackage{multicol}
\usepackage{color, colortbl}
\usepackage{pifont}%
\usepackage{array}
\usepackage{makecell}
\usepackage{balance}
\title{\LARGE \bf
RenderWorld: World Model with Self-Supervised 3D Label
}

\author{Ziyang Yan\textsuperscript{2,3*}, Wenzhen Dong\textsuperscript{4*}, Yihua Shao\textsuperscript{5*\dag},  Yuhang Lu\textsuperscript{1}, Haiyang Liu\textsuperscript{5}, Jingwen Liu\textsuperscript{5},  \\ Haozhe Wang\textsuperscript{6}, Zhe Wang\textsuperscript{4}, Yan Wang\textsuperscript{4\Letter},
Fabio Remondino\textsuperscript{2}, Yuexin Ma\textsuperscript{1\Letter}
\thanks{$^{1}$The ShanghaiTech University, Shanghai, China. $^{2}$Fondazione Bruno Kessler, Trento, Italy. 
$^{3}$The University of Trento, Trento, Italy. $^{4}$The Institute for AI Industry Research (AIR), Tsinghua University, Beijing, China. $^{5}$The University of Science and Technology Beijing, Beijing, China. $^{6}$The Hong Kong University of Science and Technology, Hong Kong, China. 
}%
\thanks{* The first three authors contributed equally.}
\thanks{\dag Project leader.}
\thanks{\Letter \raggedright{Corresponding to mayuexin@shanghaitech.edu.cn} }
}

\begin{document}
\begin{CJK}{UTF8}{gbsn}

\maketitle
\thispagestyle{empty}
\pagestyle{empty}

\begin{abstract}
End-to-end autonomous driving with vision-only is not only more cost-effective compared to LiDAR-vision fusion but also more reliable than traditional methods. To achieve a economical and robust purely visual autonomous driving system, we propose RenderWorld, a vision-only end-to-end autonomous driving framework, which generates 3D occupancy labels using a self-supervised gaussian-based Img2Occ Module, then encodes the labels by AM-VAE, and uses world model for forecasting and planning. RenderWorld employs Gaussian Splatting to represent 3D scenes and render 2D images greatly improves segmentation accuracy and reduces GPU memory consumption compared with NeRF-based methods. By applying AM-VAE to encode air and non-air separately, RenderWorld achieves more fine-grained scene element representation, leading to state-of-the-art performance in both 4D occupancy forecasting and motion planning from autoregressive world model.

\end{abstract}

\section{INTRODUCTION}

    With the wide application of autonomous driving \cite{li2022bevformer}, \cite{yang2023bevformer}, \cite{huang2021bevdet} \cite{shao2024accidentblip}, researchers gradually focus on better perception and forecasting methods \cite{yang2023one}, which are related to the decision-making ability and robustness of the system \cite{tian2024drivevlm}, \cite{cui2023drivellm}. Most current frameworks consist of perception \cite{hu2022st}, forecasting, and planning separately \cite{hu2023planning}. The most commonly used perception method is 3D target detection using vision and LIDAR fusion \cite{huang2021bevdet}, \cite{li2023bevdepth}, \cite{li2022bevformer}, allowing the model to better forecast future scenes and do motion planning. Since most 3D target detection methods \cite{zhu2021cylindrical, sadat2020perceive, wei2021perceive} are unable to obtain fine-grained information in the environment, they are non-robust in planning \cite{ma2019trafficpredict} in the subsequent model, which affects the system safety. Current perception methods primarily utilize both LiDAR \cite{zhang2022beverse}, \cite{peng2023cl3d} and cameras \cite{liu2023bevfusion}, but the high cost of LiDAR and the computational demands of multimodal fusion pose challenges to the real-time performance and robustness of autonomous driving systems.\par

     In this paper, we introduce \textbf{RenderWorld}, an autonomous driving framework for prediction and motion planning, which is trained on 3D occupancy labels generated by a Gaussian-based Img2Occ module. RenderWorld proposes an self-supervised Img2Occ module with Gaussian Splatting \cite{kerbl20233d}, trained on 2D multi-view depth and semantic images to generate 3D occupancy labels required for the world model. To enable the world model to better understand the scene represented by 3D occupancy, we propose the Air Mask Variational Autoencoder (AM-VAE) upon a vector-quantized variational autoencoder (VQ-VAE) \cite{van2017neural}. This improves the inference capability of our world model by enhancing the granularity of the scene representation.\par

     In order to verify the efficiency and reliability of RenderWorld, we evaluate the 3D occupancy generation and motion planing on NuScenes \cite{caesar2020nuscenes} separately. In summary, our contributions are mainly as follows:
     \begin{itemize}
    \item [1)] 
    We propose RenderWorld, a pure 2D autonomous driving framework that uses labeled 2D images to train a Gaussian-based occupancy prediction module (Img2Occ) for generating the 3D labels required by the world model.
    \item [2)] 
    To improve spatial representation abilities, we introduce AM-VAE, which improves forecasting and planning in world models while reducing memory consumption by separately encoding air and non-air voxels.
    \end{itemize}
\begin{figure*}[htp]
    \centering
    \includegraphics[width=12.5cm]{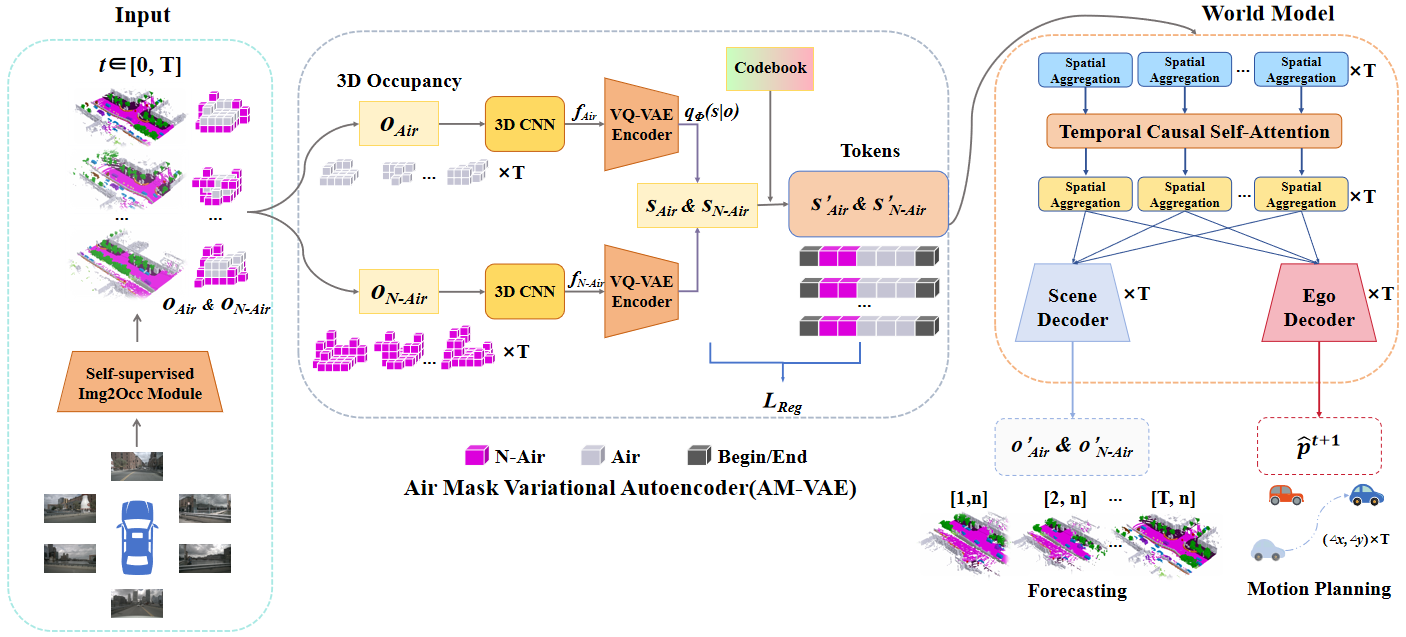}
    \caption{\textbf{General pipeline of RenderWorld.} We firstly generate the 3D occupancy labels through an Img2Occ Module (Figure~\ref{fig:gs-pipeline}). Then, using Air Mask Variational Autoencoder (AM-VAE) described in Section ~\ref{sub:AMVAE}, the separated air and non-air voxels are independently encoded into latent representations (i.e., discrete tokens). Finally, these latent representations are processed according to the specifications in Section ~\ref{sub:worldmodel}, and based on this, the voxels and trajectories are predicted, ultimately outputting the predicted occupancy and self-planning.
 }
    \label{fig:gen_pipe}
    \vspace{-2.0em}
\end{figure*}
\section{RELATED WORK}
\subsection{3D Occupancy Prediction}
3D occupancy is gaining attention as a viable alternative to LiDAR perception \cite{zhang2023occnerf}. Most previous works \cite{huang2023tri, cao2022monoscene, li2022bevformer, zhang2023occformer} utilize 3D Occupancy Ground Truth for supervision, which is challenging to annotate. With the widespread adoption of Neural Radiance Fields (NeRF) \cite{remondino2023critical, yan2023nerfbk}, some methods \cite{pan2024renderocc, zhang2023occnerf, zhao2024hybridocc, huang2024selfocc, boeder2024occflownet, liu2024let} have attempted to use 2D depth and semantic labels for training. However, using continuous implicit neural fields to predict occupancy probabilities and semantic information often leads to high memory cost \cite{yan20243dsceneeditor}. Recently, GaussianFromer \cite{huang2024gaussianformer} leverages sparse Gaussian points as a means of reducing GPU consumption to describe 3D scenes while GaussianOcc \cite{gan2024gaussianocc} utilizes a 6D pose network to eliminate the reliance on ground truth poses, but both of them suffers from a significant drop in overall segmentation accuracy. In our work, we employ an anchor-based Gaussian initialization method to gaussianize voxel fratures and represent the 3D scenes with denser Gaussian points that achieving higher segmentation accuracy while avoiding the excessive memory consumption of ray sampling in NeRF-based methods.

\subsection{World Model in Autonomous Driving}
World models \cite{ha2018world,wang2023adversarial} are often used for future frame prediction and to assist robots in making decisions \cite{sutton1990integrated}. As end-to-end autonomous driving \cite{hu2023planning}, \cite{jiang2023vad} is gradually evolving, world models are also applied for predicting future scenarios and decisions making \cite{hu2023gaia}. Unlike traditional autonomous driving approaches \cite{pomerleau1988alvinn}, \cite{bojarski2016end}, the world model approaches integrate perception, prediction and decision making. Many current approaches perform fusion of camera-LiDAR data and input into world model, which is used to forecast \cite{gu2023vip3d}, \cite{hu2021fiery}  and make motion planning \cite{huang2023gameformer}. Among them, OccWorld \cite{zheng2023occworld} proposes to utilize 3D occupancy as world model's input. However, OccWorld is less effective at utilizing pure 2D input and struggles to accurately predict future scenes due to information loss during the encoding process. Hence, we design an Img2Occ Module to convert 2D labels into 3D occupancy labels to enhance the world modeling capabilities.

\section{METHODOLOGY}
In this section, We describe the overall implementation of RenderWorld. We firstly propose an Img2Occ Module for occupancy prediction and 3D occupancy labels generation (Sec III-A). Subsequently, we introduce a module based on the Air Mask Variational Autoencoder (AM-VAE) to optimize occupancy representation and enhance data compression efficiency (Section III-B). Finally, we elaborate on how to integrate the World Model for accurate prediction of 4D scene evolution (Section III-C).

\subsection{3D Occupancy prediction with Multi-frame 2D Labels}

\begin{figure}[htp]
    \centering
    \includegraphics[width=6.5cm]{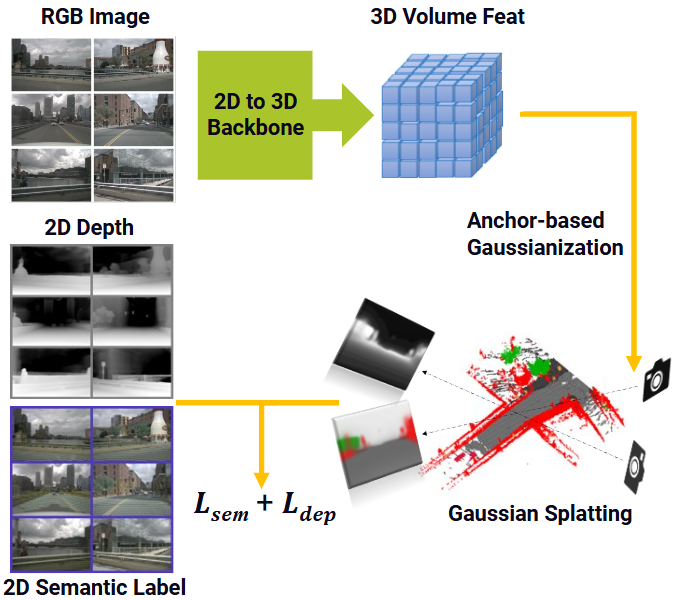}
    \caption{\textbf{Training paradigm of 2D-to-3D occupancy prediction Module.} Our proposed Img2Occ Module utilizes 2D labels to train the 3D occupancy network that allowing the model to take advantage of detailed 2D pixel-level semantics and depth supervision.}
    \label{fig:gs-pipeline}
    \vspace{-1.0em}
\end{figure}

To enable 3D semantic occupancy prediction and future 3D occupancy labels generation, we design an Img2Occ Module which is illustrated in Figure~\ref{fig:gs-pipeline}. Using images from multi-cameras $\left \{ {{{Img}_i}}\right \}_{i=1}^N  $ as inputs, we firstly extract 2D image features using a pretrained BEVStereo4D \cite{huang2022bevdet4d} backbone and Swin Transformer \cite{liu2021Swin}. Then, these 2D messages are interpolated into 3D space to produce volume features by leveraging the known intrinsic parameters  $\left \{{{I_i}}\right \}_{i=1}^N$ and extrinsic parameters $\left \{{{E_i}}\right \}_{i=1}^N$.
To project the 3D occupancy voxels onto multi-camera semantic maps, we apply Gaussian Splatting \cite{kerbl20233d}, an advanced real-time rendering pipeline. Inspired by \cite{scaffoldgs}, we initialize anchor points with a learnable scale at the center of each voxel to approximate scene occupancy. The attributes of each anchor are determined based on the relative distance and viewing direction between the camera and the anchor. This anchor set is then used to initialize a Gaussian set with semantic labels $\left \{{{G_x}}\right \}_{x=1}^N$. Each Gaussian point $x$ is then represented by a full 3D covariance matrix $\Sigma$ in world space and its center position $µ$, and the color of each point is decided by the semantic label at that point.
\begin{equation} 
G(x)=e^{-\frac{1}{2}(x-\mu)^{T} \Sigma^{-1}(x-\mu)} \end{equation}
Directly optimizing $\Sigma$ may lead to infeasible matrices as it must be positive semi-definite. To ensure the validity of $\Sigma$, it is decomposed into the scaling matrix $S$ and the rotation matrix $R$ to characterize the geometry of a 3D Gaussian ellipsoid:
\begin{equation}\mathrm{\Sigma}\ =\ RSS^TR^T\end{equation}
Then the 3D Gaussians are projected to 2D for rendering by computing the camera space covariance matrix ${\Sigma}^\prime$ :
\begin{equation}\mathrm{\Sigma}^\prime=\ JW\mathrm{\Sigma}W^TJ^T,\end{equation}
where $J$ is the Jacobian matrix of the affine approximation of the projection transformation and $W$ is the viewing transformation. The semantic / depth of each pixel can then be calculated by applying alpha blending onto sorted Gaussians:
\begin{equation}
D=\sum_{i}^{N}(d_i)a_i\prod_{j}^{i-1}{(1-a_j)},
\end{equation}
\begin{equation}
S=\sum_{i}^{N}(s_i)a_i\prod_{j}^{i-1}{(1-a_j)},
\end{equation}
where $s_i/d_i$ is the rendered  semantic / depth of a 3D Gaussian, $a_i$ is the product of an evaluated 2D Gaussian projection and its corresponding opacity.

To calculate the difference between ground truth depth and rendered depth, we utilize the Pearson correlation which can measure the distribution difference between 2D depth maps  follows the following function:
\begin{equation}L_{dep}^i = Corr(\bar{d}_i, \ {{d}_i}) = \ \frac{Cov(\bar{d}_i, \ {{d}_i)}}{Var(\bar{{d}_i}, \ {{d}_i)}},
\end{equation}
where ${\bar{d}_i}$ is the ground truth depth image  and ${{d}_i}$ is the rendered depth image.

Finally, we construct the loss function with a cross-entropy loss ${{L}_{sem}}$ for supervising semantic segmentation and ${{L}_{dep}}$ for depth supervision, the overall loss can be computed as follows:
\begin{equation}L_{i2o}^i =  L_{sem}^i + L_{dep}^i
\end{equation}
Using the well-trained checkpoint, we generate 3D occupancy labels, which are then input into the subsequent AM-VAE module.

\subsection{Air Mask Variational Autoencoder (AM-VAE)} 
\label{sub:AMVAE}

Traditional Variational Autoencoders (VAEs) fail to encode the distinct features of non-air voxels which hampers the model to represent scene elements as fine-grained level. To address this issue, we introduce the Air Mask Variational Autoencoder (AM-VAE), a novel VAE involves training two distinct Vector Quantized Variational Autoencoders (VQ-VAE) \cite{van2017neural} to encode and decode air and non-air occupancy voxels separatly.

Assuming \( o \) represents the input occupancy representation, and \( o_{Air} \) and \( o_{N-Air} \) represent the air and non-air voxels. We first utilize a 3D convolutional neural network to encode the occupancy data, with the output being a continuous latent space representation denoted as \( f \). The encoder \( q_\phi(s|o) \) maps the input \( f \) to the latent space \( s \). Then, we use two latent variables \( s_{Air} \) and \( s_{N-Air} \) to represent the air and non-air voxels, respectively:
\begin{equation}
s_{Air} \sim q_\phi(s_{Air}|o_{Air}), \quad s_{N-Air} \sim q_\phi(s_{N-Air}|o_{N-Air})
\end{equation}

Each encoded latent variable \( s_{Air} \) or \( s_{N-Air} \) uses learnable codebook \( C_{Air} \) or \( C_{N-Air} \) to obtain discrete token, which is then replaced by the most similar codebook entry before being fed into the decoder. This process is represented as:

\begin{equation}
\begin{aligned} 
s'_{Air} &= \arg\min_{c_{Air} \in C_{Air}} \| s_{Air} - c_{Air} \|, \\
s'_{N-Air} &= \arg\min_{c_{N-Air} \in C_{N-Air}} \| s_{N-Air} - c_{N-Air} \|
\end{aligned} 
\end{equation}

Then, the decoder \( p_\theta(o|s) \) reconstructs the input occupancy from the quantized latent variables \( s'_{Air} \) and \( s'_{N-Air} \):

\begin{equation}
\hat{o}_{Air} = p_\theta(o_{Air}|s'_{Air}), \quad \hat{o}_{N-Air} = p_\theta(o_{N-Air}|s'_{N-Air})
\end{equation}

To facilitate the separation of air and non-air elements within the occupancy representation, we denote \( M \) as the set of non-air categories. Then the indicator function for air and non-air in the modified occupancy can be defined as follows:


\begin{equation}
I_M(o) = 
  \begin{cases} 
    1 & \mathit{if } o \in M, \\
    0 & \mathit{otherwise.}
  \end{cases}
\end{equation}

The modified air occupancy \( o'_{Air} \) and non-air occupancy \( o'_{N-Air} \) are given by the following equations:

\begin{equation}
\begin{aligned} 
o'_{Air} &= (1 - I_M(o)) \cdot o_{Air},\\
\quad o'_{N-Air} &= I_M(o) \cdot o_{N-Air} + (1 - I_M(o)) \cdot o_{Air}
\end{aligned} 
\end{equation}

To reconstruct the original occupancy representation, we use a \( \mathit{mask} = (\hat{o}_{Air} \neq 0) \) to distinguish areas filled only with air. Then the reconstructed occupancy \( \hat{o} \) combines the air and non-air components as follows:
\begin{equation}
\hat{o} = \hat{o}_{Air} \cdot \mathit{mask} + \hat{o}_{N-Air} \cdot (1 - \mathit{mask})
\end{equation}

We then build the loss function ${{L}_{VAE}}$  for training the AM-VAE with reconstruction loss ${{L}_{Recon}}$ and commitment loss ${{L}_{Reg}}$:

\begin{equation}  \label{eqn: reconloss}
\begin{aligned}
\mathit{L_{Recon}} ={}& \mathbb{E}_{q_\phi(s_{Air}|o_{Air})} [\log p_\theta(o_{Air}|s'_{Air})] \\  
    & + \mathbb{E}_{q_\phi(s_{N-Air}|o_{N-Air})} [\log p_\theta(o_{N-Air}|s'_{N-Air})],
\end{aligned}  
\end{equation}
\begin{equation}  \label{eqn: reconloss}
\begin{aligned}
\mathit{L_{Com}} = \| s_{Air} - s'_{Air} \|^2 + \| s_{N-Air} - s'_{N-Air} \|^2 ,
\end{aligned}  
\end{equation} 
\begin{equation}  \label{eqn: vaeloss}
\begin{aligned}  
\mathit{L_{VAE}} =\mathit{L_{Recon}}  + \beta \mathit{L_{Com}}
\end{aligned}  
\end{equation}

AM-VAE utilizes separate codebooks for air and non-air voxels within a unified encoder-decoder setup. This method effectively captures the unique features of each voxel type, thereby improving both reconstruction accuracy and generalization potential.

\subsection{World Model}
\label{sub:worldmodel}
By applying a world model in autonomous driving to encode 3D scenes into high-level tokens, our framework can effectively capture environmental complexity, enabling accurate autoregressive anticipation of future scenarios and vehicle decisions.

 Inspired by OccWorld~\cite{ratliff2006maximum}, we use a 3D occupancy to represent the scene and employ a self-supervised tokenizer to derive high-level scene tokens $\mathbf{T}$, and encode the spatial position of vehicles by aggregating the vehicle token $\mathbf{z}_0$. The world model is defined as $w$ based on the current timestamp $T$ and the number of historical frames $t$, then we establish the prediction with the following formula:
\begin{equation}
w (\mathbf{T}^T, \cdots, \mathbf{T}^{T-t}) = \mathbf{T}^{T+1},
\end{equation}
where $\mathbf{T}^{T+1}$ represents the scene tokens at the next time step.

At the same time, a temporal generative transformer architecture is adopted to effectively predict the future scene. It firstly processes scene tokens through spatial aggregation and downsampling, and then generates a hierarchical set of tokens $\{\mathbf{T}_0, \cdots, \mathbf{T}_K \}$. So as to predict the future at different spatial scales, we take multiple sub-world models $w = \{w_0, \cdots, w_K \}$ to achieve it and each sub-model $w_i$ applies temporal attention to the tokens at each position $j$ using the following formula:
\begin{equation}
\hat{\mathbf{z}}^{T+1}_{j,i}  = \text{TA} (\mathbf{z}^{T}_{j,i}, \cdots, \mathbf{z}^{T-t}_{j,i}),
\end{equation}
where TA represents masked temporal attention, which predicts future tokens from influencing previous tokens. $\mathbf{z}^{t}_{j,i} \in \mathbf{T}^{t}_{i}$ denotes the $j$-th world token at scale $i$ and timestamp $t$. 

In the prediction module, we firstly utilize a self-supervised tokenizer $e$ to convert the 3D scene into high-level scene tokens $\mathbf{T}$, and a vehicle token $\mathbf{z}_0$ to encode the spatial position of the vehicle. After predicting the future scene tokens, a scene decoder $d$ is applied to decode the predicted 3D occupancy $\hat{\mathbf{y}}^{T+1} = d(\hat{\mathbf{z}}^{T+1})$, and learn a vehicle decoder $d_{ego}$ which is for generating the vehicle displacement that relative to the current frame $\hat{p}^{T+1} = d_{ego}(\hat{z}^{T+1}_0)$. The prediction module provides decision support for trajectory optimization of the autonomous driving system by generating continuous predictions of future vehicle displacements and scenario changes, ensuring safe and adaptive path planning.

We have implemented a two-stage training strategy to effectively train our prediction module. In the first phase, we train the scene tokenizer $e$ and the decoder $d$ using a 3D occupancy loss: 
\begin{equation}\label{eqn: loss_world1}
\begin{aligned}
L_{e,d} = L_{soft}(d(e(\mathbf{y})), \mathbf{y}) + \lambda_1 \cdot L_{lovasz}(d(e(\mathbf{y})), \mathbf{y}),
\end{aligned}
\end{equation}
where $L_{soft}$ denotes the softmax loss and $L_{lovasz}$ represents the Lovasz-softmax loss. The term $\lambda_1$ serves as a balancing factor between them.

Then we use the learned scene tokenizer $e$ to obtain the scene tokens $\mathbf{z}$ for all frames and constrain the difference between the predicted tokens $\hat{\mathbf{z}}$ and $\mathbf{z}$. And a softmax loss is used to enforce the correct classification of $\hat{\mathbf{z}}$ to the correct code in the codebook $\mathbf{C}$. For the vehicle token, we simultaneously learn the vehicle decoder $d_{ego}$ and apply an L2 loss on the predicted displacement $\hat{p} = d_{ego}(\hat{\mathbf{z}}_0)$ and the ground truth displacement $\mathbf{p}$. The overall loss in phase two can be formulated as follows:
\begin{equation}\label{eqn: loss_world}
\begin{aligned}
  L_{w,d_{ego}} = \sum_{t=1}^T (\sum_{j=1}^{M_0}  L_{soft} (\hat{\mathbf{z}}_{j,0}^t , \mathbf{C}(\mathbf{z}_{j,0}^t) \\
  + \  \lambda_2 \  L_{L2} (d_{ego}(\hat{\mathbf{z}}^t_0), \mathbf{p}^t)),
\end{aligned}
\end{equation}
where $T$ and $M_0$ are the number of frames and the number of spatial tokens at the original scale, respectively. $\mathbf{C}(\cdot)$ denotes the index of the corresponding code in the codebook $\mathbf{C}$. $L_{L2}$ measures the L2 difference between the two trajectories. 


\begin{table*}[htbp]

\centering
    \resizebox{\textwidth}{!}{
    \begin{tabular}{l|c|c|c c c c c c c c c c c c c c c c c c}
    \toprule
    Methods & GT & mIoU $\uparrow$ & \rotatebox{90}{Others} & \rotatebox{90}{barrier} & \rotatebox{90}{bicycle} & \rotatebox{90}{bus} & \rotatebox{90}{car}
 & \rotatebox{90}{cons. veh} & \rotatebox{90}{motorcycle} & \rotatebox{90}{pedestrian} & \rotatebox{90}{traffic cone} & \rotatebox{90}{trailer} & \rotatebox{90}{truck} & \rotatebox{90}{dri. sur} & \rotatebox{90}{other flat} & \rotatebox{90}{sidewalk} & \rotatebox{90}{terrain} & \rotatebox{90}{manmade}  & \rotatebox{90}{vegetation}  \\
    \midrule
    TPVFormer \cite{huang2023tri} & 3D & 27.83 & 7.22 & 38.90 & 13.67 & \textbf{40.78}& \textbf{45.90} & \textbf{17.23}& 19.99 & 18.85& 14.30& \textbf{26.69}& \textbf{34.17}& 55.65& 35.47 & 37.55& 30.70& 19.40& 16.78\\
    BEVFormer \cite{li2022bevformer} & 3D & 26.88 & 5.03 & 38.79 & 9.98 & 34.41 & 41.09 & 13.24 & 16.50 & 18.15 & 17.83 & 18.66 & 27.70 & 48.95 & 27.73 & 29.08 & 25.38 & 15.41 & 14.46 \\
    OccFormer \cite{zhang2023occformer} & 3D & 21.93 & 5.94 & 30.29 & 12.32 & 34.40 & 39.17 & 14.44 & 16.45 & 17.22 & 9.27 & 13.90 & 26.36 & 50.99 & 30.96 & 34.66 & 22.73 & 6.76 & 6.97 \\

    CTF-Occ \cite{tian2024occ3d} & 3D & \textbf{28.53} & \textbf{8.09} & \textbf{39.33} & \textbf{20.56} & 38.29 &  42.24 & 16.93 & \textbf{24.52} & \textbf{22.72} & \textbf{21.05} & 22.98 & 31.11 & 53.33 & 33.84 & 37.98 & 33.23 & 20.79 & 18.0 \\
    \midrule
    RenderOcc \cite{pan2024renderocc} & 2D & 23.93 & 5.69 & 27.56 & 14.36 & 19.91 & 20.56 & 11.96 & 12.42 & 12.14 & 14.34 & 20.81 & 18.94 & 68.85 & 33.35 & 42.01 & 43.94 & 17.36 & 22.61 \\
    SurroundOcc \cite{wei2023surroundocc} & 2D & 20.30 & -
    & 20.59 & 11.68 & 28.06 & 30.86 & 10.70 & 15.14 & 14.09 & 12.06 & 14.38 & 22.26 & 37.29 & 23.70 & 24.49 & 22.77 & 14.89 & 21.86 \\
    GaussianFormer \cite{huang2024gaussianformer} & 2D & 19.10 & - & 19.52 & 11.26 & 26.11 & 29.78 & 10.47 & 13.83 & 12.58 & 8.67 & 12.74 & 21.57 & 39.63 & 23.28 & 24.46 & 22.99 & 9.59 & 19.12 \\
    GaussianOcc \cite{gan2024gaussianocc} & 2D & 9.94 & - & 1.79 & 5.82 & 14.58 & 13.55 & 1.30 & 
  2.82 & 7.95 &  9.76 & 0.56 & 9.61 & 44.59 & - & 20.10 & 17.58 & 8.61 & 10.29\\
    OccNeRF \cite{zhang2023occnerf} & 2D & 9.53  & - & 0.83 & 0.82 & 5.13 & 12.49 & 3.50 & 
 0.23 & 3.10 & 1.84 & 0.52 & 3.90 & 52.62 & - & 20.81 & 24.75 & 18.45 & 13.19\\
     SelfOcc \cite{huang2024selfocc} & 2D & 9.30 & 0.00 & 0.15 & 0.66 & 5.46 & 12.54 & 0.00 & 0.80 & 2.10 & 0.00 & 0.00 & 8.25 & 55.49 & 0.00 & 26.30 & 26.54 & 14.22 & 5.60 \\
    \textbf{RenderWorld (Ours)} & 2D & 27.87 & 6.83 & 32.54 & 7.44 & 21.15 & 29.92 & 16.68 & 11.43 & 17.45 & 16.48 & 24.02 & 27.86 & \textbf{75.05} & \textbf{36.82} & \textbf{50.12} & \textbf{53.04} & \textbf{22.75} & \textbf{24.23} \\
    \bottomrule
\end{tabular}}

\caption{\textbf{3D Occupancy prediction performance on the Occ3D-nuScenes validation set. } Our method outperforms state-of-the-art methods, particularly excelling in environment-related categories  (i.e. terrain, vegetation.).}
\label{tab:semantic_result}
\vspace{-1.0em}
\end{table*}
\section{EXPERIMENTS}
In this section we evaluate the performance of RenderWorld using NuScenes \cite{caesar2020nuscenes} dataset. We also performed extensive ablation experiments on the same dataset - as reported in sub-section C - to deeper understand the proposed approach.

\subsection{Experimental Setup}
We adopt NuScenes as our evaluation dataset. NuScenes is a large-scale autonomous driving dataset that includes 700 scenes for training, 150 scenes for validation, and 150 scenes for testing, totaling approximately 40,000 frames across 17 classes. For self-supervised training, we generate ground truth depths and 2D segmentation ground truths by projecting LiDAR point clouds with their 3D segmentation labels onto corresponding 2D views. During the semantic occupancy prediction, each sample covers a range of [x:(-40 m, 40 m), y:(-40 m, 40 m), z:(-1.0 m, 5.4 m)] with a voxel size of 0.4 m. The evaluation experiments of our model are conducted on the 150 validation sets with one NVIDIA A30 GPU. 

\begin{table*}[htbp]
\resizebox{\textwidth}{!}{
\centering
\begin{tabular}{l|cc|cccc|cccc|c}
\toprule
Method & Input & Aux. Sup. &
\multicolumn{4}{c|}{mIoU $\uparrow$} & 
\multicolumn{4}{c|}{IoU $\uparrow$} & \\
&& & 1s & 2s & 3s & Avg.  & 1s & 2s & 3s & Avg. &Memory\\
\midrule
Copy\&Paste & 3D-Occ & None & 14.91 & 10.54 & 8.52 & 11.33 & 24.47 & 19.77 & 17.31 & 20.52  & -\\

OccWorld (Original)~\cite{zheng2023occworld} & 3D-Occ & None  & 25.78 & 15.14 & 10.51 & 17.14   & 34.63 & 25.07 & 20.18 & 26.63  & 13500M\\

\textbf{RenderWorld(Ours)} & 3D-Occ & None  & \textbf{28.69} & \textbf{18.89} & \textbf{14.83} & \textbf{20.80}   & \textbf{37.74} & \textbf{28.41} & \textbf{24.08} & \textbf{30.08}  & \textbf{13000M}\\

\hline

TPVFormer~\cite{huang2023tri}+Lidar+OccWorld-T~\cite{zheng2023occworld} & Camera & Semantic LiDAR  & \textbf{4.68} &  \textbf{3.36} & \textbf{2.63} & \textbf{3.56}  & 9.32 & 8.23 & 7.47 & 8.34   & 15000M\\
TPVFormer~\cite{huang2023tri}+SelfOcc~\cite{huang2024selfocc}+OccWorld-S~\cite{zheng2023occworld} & Camera & None & 0.28 & 0.26 & 0.24 & 0.26 &  5.05 & 5.01 & 4.95 & 5.00  & 15000M\\

\textbf{RenderWorld(Ours)} & Camera & None  & 2.83 & 2.55 & 2.37 & 2.58   & \textbf{14.61} & \textbf{13.61} & \textbf{12.98} & \textbf{13.73}   & \textbf{14400M}\\
\bottomrule
\end{tabular}}%

\caption{\textbf{4D occupancy forecasting performance.}
Aux. Sup. denotes auxiliary supervision apart from the ego trajectory.
Avg. denotes the average performance of that in 1s, 2s, and 3s.
}
\label{occforecast}
\vspace{-1.0em}
\end{table*}

\begin{table*}[htbp]
\setlength{\tabcolsep}{0.008\linewidth}
\centering
\begin{tabular}{l|c|c|cccc|cccc}
\toprule
\multirow{2}{*}{Method} & \multirow{2}{*}{Input} & \multirow{2}{*}{Aux. Sup.} &
\multicolumn{4}{c|}{L2 (m) $\downarrow$} &
\multicolumn{4}{c}{Collision Rate (\%) $\downarrow$}  \\
&& & 1s & 2s & 3s & Avg.& 1s & 2s & 3s & Avg. \\
\midrule
IL~\cite{ratliff2006maximum} & LiDAR & None  & 0.44 & 1.15 & 2.47 & 1.35  & 0.08 & 0.27 & 1.95 & 0.77   \\
NMP~\cite{zeng2019end} & LiDAR & Box \& Motion & 0.53 & 1.25 & 2.67 & 1.48 & \textbf{0.04} & \underline{0.12} & \underline{0.87} & 0.34  \\
FF~\cite{hu2021safe} & LiDAR & Freespace  & 0.55 & 1.20 & 2.54 & 1.43 & 0.06 & 0.17 & 1.07 & 0.43 \\
EO~\cite{khurana2022differentiable} & LiDAR & Freespace  & 0.67 & 1.36 & 2.78 & 1.60 & \textbf{0.04} & \textbf{0.09} & 0.88 & \underline{0.33}\\
\midrule
ST-P3~\cite{hu2022st} & Camera & Map \& Box \& Depth & 1.33 & 2.11 & 2.90 & 2.11   & 0.23 & 0.62 & 1.27 & 0.71\\
UniAD~\cite{hu2023planning} & Camera & { \footnotesize Map \& Box \& Motion \& Tracklets \& Occ}  & 0.48 & \underline{0.96} & \textbf{1.65} & \textbf{1.03} & 0.05 & 0.17 & \textbf{0.71} & \textbf{0.31} \\
VAD-Tiny~\cite{jiang2023vad}  & Camera & Map \& Box \& Motion  & 0.60 & 1.23 & 2.06 & 1.30 & 0.31 & 0.53 & 1.33 & 0.72  \\
VAD-Base~\cite{jiang2023vad} & Camera & Map \& Box \& Motion & 0.54 & 1.15 & 1.98 & 1.22 & \textbf{0.04} & 0.39 & 1.17 & 0.53 \\
OccNet~\cite{tong2023scene} & Camera & 3D-Occ \& Map \& Box & 1.29 & 2.13 & 2.99 & 2.14 & 0.21 & 0.59 & 1.37 & 0.72  \\
OccWorld-T~\cite{zheng2023occworld} & Camera & Semantic LiDAR & 0.54 & 1.36 & 2.66 & 1.52 & 0.12 & 0.40 & 1.59 & 0.70 \\
OccWorld-S~\cite{zheng2023occworld} & Camera & None & 0.67 & 1.69 & 3.13 & 1.83 & 0.19 & 1.28 & 4.59 & 2.02\\
\textbf{RenderWorld(Ours)} & Camera & None & 0.48 & 1.30 & 2.67 & 1.48 & 0.14 & 0.55 & 2.23 & 0.97  \\

\midrule

OccNet~\cite{tong2023scene} & 3D-Occ & Map \& Box & 1.29 & 2.31 & 2.98 & 2.25 & 0.20 & 0.56 & 1.30 & 0.69  \\
OccWorld~\cite{zheng2023occworld} & 3D-Occ & None & \underline{0.43} & 1.08 & 1.99 & \underline{1.17} & 0.07 & 0.38 & 1.35 & 0.60  \\
\textbf{RenderWorld(Ours)} & 3D-Occ & None & \textbf{0.35} & \textbf{0.91} & \underline{1.84} & \textbf{1.03} & \underline{0.05} & 0.40 & 1.39 & 0.61  \\
\bottomrule
\end{tabular}%

\caption{\textbf{Motion planning performance.} 
Aux.Sup.denotes auxiliary supervision apart from the ego trajectory.
}
\label{tab:sota-plan}
\vspace{-3.0em}
\end{table*}

\subsection{Main Result}
\textbf{3D semantic occupancy prediction:} To demonstrate the performance of our model, we compare it against 10 occupancy prediction models, which are the existing common models evaluated on the NuScenes dataset. The results in Table \ref{tab:semantic_result} indicate that RenderWorld outperforms most state-of-the-art occupancy prediction methods in mIoU, ranking second overall, and only surpassed by CTF-OCC \cite{tian2024occ3d}, which uses 3D occupancy GT as input. Furthermore, our method achieves outstanding performance in vehicle segmentation, including trailers, construction vehicles, trucks, etc and surpasses all other methods in segmenting various environmental terrains, such as vegetation, sidewalk etc. This is due to the 3D Gaussian representation, which effectively leverages the sparsity and object diversity in driving scenes, scaling with flexible location and covariance properties \cite{huang2024gaussianformer}.\par

\begin{figure*}[htp]
    \centering
    \includegraphics[width=15cm]{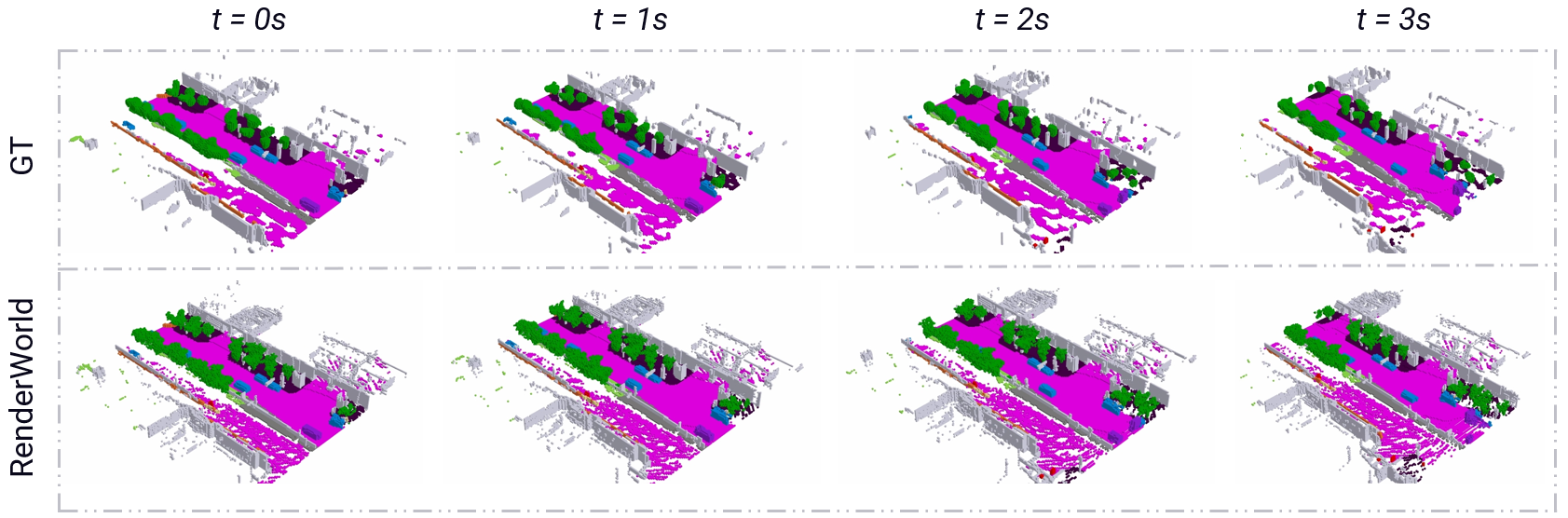}
    \caption{\textbf{Visualization of the forecasting and planning results of RenderWorld.} }
    \label{fig:world-vis}
\end{figure*}

\begin{table*}[t]
\setlength{\tabcolsep}{0.016\linewidth}
\centering
\begin{tabular}{c|cccc|cccc}
\toprule
 \multirow{2}{*}{Setting}  &
\multicolumn{4}{c|}{Forecasting mIoU (\%) $\uparrow$} & 
\multicolumn{4}{c}{Planning L2 (m) $\downarrow$}  \\
& 1s & 2s & 3s & Avg. & 1s & 2s & 3s & Avg. \\
\midrule
($50^2$, 128, 512)  & \textbf{28.69} & \textbf{18.89} & \textbf{14.83} & \textbf{20.80} & \textbf{0.35} & 0.91 & 1.84 & 1.03  
 \\
($50^2$, 128, 256)  & 27.16 & 18.09 & 14.45 & 19.90 & \textbf{0.35} & \textbf{0.87} & \textbf{1.81} & \textbf{1.01}   
 \\
 ($50^2$, 128, 1024)  & 26.34 &18.37 &14.97 &19.89  &
 0.39 &1.05 &2.16 &1.20   \\
 ($25^2$, 256, 512)  & 15.15 &12.01 &9.56 & 12.24  &
 3.21 &5.98 &8.92 &6.04  \\
 ($100^2$, 128, 512) & 21.68 &15.07 &11.67 & 16.14  &
 0.45 &1.29 &2.28 &1.34  \\
\bottomrule
\end{tabular}%
\caption{Effect of different hyperparameters for the scene tokenizer, the setting denotes latent spatial resolution, latent channel dimension, and the codebook size respectively.
}
\label{tab:hyper}
\vspace{-3.0em}
\end{table*}

\textbf{4D occupancy forecasting:}
We evaluated the 4D occupancy forecasting performance under several settings as shown in Table~\ref{occforecast}

In order to capture finer-grained scene features and provide precise information for predictions, air-separation technique is applied to prioritize crucial non-air components in the scene, boosting prediction accuracy and computational efficiency. The results show that RenderWorld can generate non-trivial future 3D occupancy, with results far superior to OccWorld and Copy\&Paste, which indicates that our model learns the underlying scene evolution.

\textbf{Motion planning:}
As shown in Table ~\ref{tab:sota-plan}, We compare the motion planning performance between the proposed RenderWorld and state-of-the-art methods, and evaluate our model across various settings used in the 4D occupancy forecasting task. RenderWorld outperforms all compared methods in L2 metrics when takes 3D occupancy as input. Without any auxiliary support, our approach also achieves competitive results in collision rate and even outperforms OccWorld-S in when only uses 2D as input.

\subsection{Ablation Study}
With the aim of showing the effectiveness of our innovative modules, we conduct three ablation studies and the results are shown in Table ~\ref{tab:efficiency_result}, ~\ref{tab:hyper} and Table~\ref{ablation}:

\textbf{Efficiency comparisons among different representations:} In Table \ref{tab:efficiency_result}, we present the efficiency comparisons of various representations, highlighting that 3D Gaussian surpasses all competitors with significantly reduced memory usage. Leveraging its explicit representation, this approach assigns specific semantic data to individual 3D Gaussians, facilitating the transition from scene depiction to occupancy forecasts. This method also circumvents the high memory usage linked to the ray initialization step in NeRF-based techniques. Although our method has higher GPU memory overhead compared to GaussianFormer, it avoids the trade-off of reducing the number of Gaussian points to save memory, but leading to a loss of semantic information. 

\begin{table}[htbp]
\centering
    \resizebox{235pt}{!}{
    \begin{tabular}{l|c|c|c}
    \toprule
    Methods & Query Form & Query Resolution & Memory  \\
    \midrule
    BEVFormer \cite{li2022bevformer} & 2D BEV & $200\times 200$  & 25100 M \\
    
    TPVFormer \cite{huang2023tri} & \makecell[c]{2D Tri-Plane}  & $\makecell[c]{200\times \\ (200+16+16)}$& 29000 M \\
    PanoOcc \cite{wang2024panoocc} & 3D Voxel & $100\times100\times16$ & 35000 M \\
    Fb-occ \cite{li2023fb} & \makecell[c]{3D Voxel \&\\ 2D BEV}  & \makecell[c]{$200\times200\times16$  \\ \& $200\times200$} & 31000 M \\
    OctreeOcc \cite{lu2023octreeocc} & Octree Query & 91200 & 26500 M \\
    OccNeRF \cite{zhang2023occnerf} & 3D Voxel & $200\times200\times16$ & 79000 M \\
    RenderOcc \cite{pan2024renderocc} & 3D Voxel & $200\times200\times16$ & 23000 M \\
    GaussianFormer \cite{huang2024gaussianformer} & 3D Gaussian & 144000 & \textbf{6229 M} \\
    \midrule
    \textbf{RenderWorld (Ours)} & 3D Gaussian & 640000 & 14400M \\
    \bottomrule
\end{tabular}}
\caption{Efficiency comparison on the nuScenes dataset. The results show that 3D Gaussian significantly reduces memory usage compared to other methods with other representations.}
\label{tab:efficiency_result}
\vspace{-2.0em}
\end{table}

\textbf{Analysis of the scene tokenizer.} Table~\ref{tab:hyper} demonstrates the impact of different hyperparameter settings on the performance of scene tokenizer, our parameters are designed like OccWorld. Larger spatial resolutions can enhance reconstruction accuracy but hinder prediction and planning, because limited token capacity for learning high-level concepts complicates future predictions~\cite{zheng2023occworld}. Codebook sizes exceeding 512 lead to overfitting, while smaller sizes or resolutions compromise scene representation accuracy.

\textbf{Effects of Mask module and VAE Separation Operation.} Table~\ref{ablation} presents the ablation study about our AM-VAE module (Separate VAE refers to dividing the potential space of the VAE into air and non-air portions). The introduction of the Air-Mask module leads to a performance improvement, achieving an mIoU of 37.68\%. When applying both the Mask module and VAE separation operation together, the performance can noticeably reach to 40.25\%. This indicates that our proposed Mask and VAE Separation operation offers substantial advantages in enhancing model reconstruction accuracy and reducing positional errors. Overall, the ablation study underscores the effectiveness of the proposed enhancements, especially the air separation strategy, in substantially boosting the performance of the RenderWorld framework.


\begin{table}[htbp]
\centering
\renewcommand{\arraystretch}{1.0}

\begin{tabular}{cc|c}
\hline
Air-Mask & \makecell{Separate VAE} & mIoU \\ \hline
- & - & 35.13  \\
$\surd$ & - & 37.68  \\
- & $\surd$ & 35.42 \\
$\surd$ & $\surd$ & \textbf{40.25}  \\ \hline
\end{tabular}
\caption{Ablation studies of air mask and vae separation. Each value indicates the performance on the validation dataset.
}
\label{ablation}
\vspace{-3.0em}
\end{table}
\section{CONCLUSIONS}
In this paper, we introduce RenderWorld, a end-to-end autonomous driving framework trained on 3D occupancy labels generated by a Gaussian-based Img2Occ module and use world model for forecasting and motion planning. By leveraging Gaussian Splatting and AM-VAE, we successfully reduce GPU memory usage by at least half in 3D occupancy label generation compared to NeRF-based approaches, while simultaneously attaining minimal memory requirements in 4D occupancy forecasting. Experimental results demonstrate that our approach can achieve state-of-the-art performance in semantic segmentation of 3D occupancy, 4D occupancy forecasting with 2D input and motion planning among all input types. Our work offers a valuable contribution to the autonomous driving community, enhancing real-time, efficient robot perception, forecasting and motion planning.

\section*{Acknowledgements}
This work was supported by NSFC (No.62206173), Shanghai Frontiers Science Center of Human-centered Artificial Intelligence (ShangHAI), MoE Key Laboratory of Intelligent Perception and Human-Machine Collaboration (KLIP-HuMaCo).

\addtolength{\textheight}{-0cm}   
\balance
\bibliographystyle{IEEEtran}
\bibliography{ref}  
\end{CJK}
\end{document}